\documentclass[sigconf]{acmart}

\AtBeginDocument{%
  \providecommand\BibTeX{{%
    \normalfont B\kern-0.5em{\scshape i\kern-0.25em b}\kern-0.8em\TeX}}}
    
\setcopyright{rightsretained}

\theoremstyle{definition}

\newcounter{quotecount}

\begin{document}

\copyrightyear{2020}
\acmYear{2020}
\acmConference[FAT* '20]{Conference on Fairness, Accountability, and Transparency}{January 27--30, 2020}{Barcelona, Spain}
\acmBooktitle{Conference on Fairness, Accountability, and Transparency (FAT* '20), January 27--30, 2020, Barcelona, Spain}\acmDOI{10.1145/3351095.3373157}
\acmISBN{978-1-4503-6936-7/20/02}

\title{Value-laden Disciplinary Shifts in Machine Learning}

\author{Ravit Dotan}
\affiliation{
  \department{Philosophy Department}
  \institution{University of California, Berkeley}
}
\email{ravit.dotan@berkeley.edu}
\author{Smitha Milli}
\affiliation{%
  \department{Computer Science Department}
  \institution{University of California, Berkeley}
}
\email{smilli@berkeley.edu}

\begin{abstract}
As machine learning models are increasingly used for high-stakes decision making, scholars have sought to intervene to ensure that such models do not encode undesirable social and political values. However, little attention thus far has been given to how values influence the machine learning \emph{discipline} as a whole. How do values influence what the discipline focuses on and the way it develops? If undesirable values are at play at the level of the discipline, then intervening on particular models will not suffice to address the problem. Instead, interventions at the disciplinary-level are required.

This paper analyzes the discipline of machine learning through the lens of philosophy of science. We develop a conceptual framework to evaluate the process through which types of machine learning models (e.g. neural networks, support vector machines, graphical models) become predominant. The rise and fall of model-types is often framed as objective progress. However, such disciplinary shifts are more nuanced. First, we argue that the rise of a model-type is self-reinforcing--it influences the way model-types are evaluated. For example, the rise of deep learning was entangled with a greater focus on evaluations in compute-rich and data-rich environments. Second, the way model-types are evaluated encodes loaded social and political values. For example, a greater focus on evaluations in compute-rich and data-rich environments encodes values about centralization of power, privacy, and environmental concerns.
\end{abstract}



\maketitle

\section{Introduction}
Given the increased use of machine learning models for high-stakes decision making, a growing body of work aims to understand and intervene in the social values embedded within machine learning models. For example, to understand whether such models exacerbate existing racial, gender, or other disparities between groups and to intervene to mitigate such disparities \citep{buolamwini2018gender, chen2019fairness,chouldechova2017fair,friedler2019comparative,celis2019classification,hardt2016equality}.

We ask a complementary question: how are societal, political, environmental, and other values embedded within the machine learning \emph{discipline}? In particular, how do values influence what the discipline focuses on and the way it develops? It is important to think about these questions because when undesirable values are at play at the level of the \emph{discipline}, intervening on particular models will not suffice to address the problem. Rather, the correct intervention must also be posed at the level of the discipline.

The focus of this paper is how values shape the development of the discipline over time. We argue that major disciplinary shifts within the machine learning discipline are not (and cannot be) ``objective'' processes --- instead, they are \emph{value-laden}. This is a consequential distinction. Certain values are still implicit even when disciplinary shifts are incorrectly seen as objective progress. However, because they are hidden they are simply accepted as a default. In order to make any intentional choices about values, we must first recognize that they exist and are at work.

Our argument proceeds in three parts.

\textit{Section \ref{sec:model-types}:} First, we give a conceptual, descriptive framework for progress in machine learning. We argue that ``model-types" within machine learning, e.g. deep learning, graphical models, support vector machines, guide and organize research activities in machine learning. We point out similarities and differences between model types and traditional concepts from philosophy of science - Kuhn's paradigms and Lakatos's research programmes.

\textit{Section \ref{sec:imagenet}:} Second, we argue that the rise of a model-type is self-reinforcing: it comes hand in hand with the rise of corresponding criteria for evaluating model-types. We use the recent rise of deep learning as a case study to illustrate this point. We visit a commonly-cited cause for the rise of deep learning: its success in the 2012 ImageNet challenge. However, we argue that ImageNet not only triggered a shift to deep learning, but also a shift to evaluating models in the environments that deep learning performs best in, namely, compute-rich and data-rich environments.

\textit{Section \ref{sec:values}:} Third, we argue that criteria used to evaluate model-types encode loaded social and political values. We again illustrate the point using deep learning as a case study. Deep learning performs better when evaluated in compute-rich and data-rich environments, and typically better when evaluated on predictive accuracy (as compared to other evaluations around robustness or interpretability). This kind of evaluation furthers certain values, such as centralization of power, while hindering other values, such as environmental sustainability and privacy. Therefore, the rise of deep learning is not straightforwardly ``objective" but, rather, is a value-laden process.

\paragraph{Our contributions.} We have written our paper to be of interest to both philosophers of science and machine learning researchers, and our contributions are slightly different for each group. For both groups, we give a more nuanced account of disciplinary shifts in machine learning, and highlight ways in which values shape the discipline. For machine learning researchers, we hope the explicit exposition helps to bring into awareness and shift some of the values present in the discipline. In addition, for philosophers of science, our descriptive account of machine learning is a conceptual contribution on its own. First, since to our knowledge this is the first work analyzing machine learning as a discipline from a philosophy of science perspective\footnote{Related work in philosophy of science includes work on how philosophy of science and machine learning can illuminate one another on topics like inductivism, the logic of discovery, and scientific realism. For examples, see \citep{Fricke2015,Kitchin2014,Thagard1990,Korb2004,Williamson2004,Williamson2009,Gillies1996,Bensusan2000,Bergadano1993,Corfield2010}}, we hope that our framework provides a starting point for others to further analyze the growing discipline of machine learning. Second, considering the case of machine learning can be helpful in reexamining traditional concepts in philosophy of science. In particular, our analysis suggests that it may be possible to reproduced traditional concepts and puzzles from the philosophy of science without the normal focus on theories and hypotheses.

\section{Model types as organizing and guiding research} \label{sec:model-types}
Machine learning models can be grouped into different types based on the ways that they extract patterns from data. Deep learning models (neural networks), graphical models, decision trees, and support vector machines, are all examples of model types. We argue that model types are more than just a technical apparatus. Model types guide the research agenda of machine learning practitioners who are committed to them, and when many people are committed to the same model type the discipline and the resources available to practitioners change. 

We start this section by explaining what it means to be committed to a model type. We then explain how this commitment influences research agendas and the discipline. Last, we point out analogies and differences between the function of model types in machine learning and paradigms and research programmes in natural sciences. These will later be used to argue that comparison between model types is not an objective process.

\subsection{Commitment to model-types}
Researchers often associate their work with a specific model-type. For example, they sometimes identify themselves as working on specifically ``deep learning" or ``graphical models". This reflects the fact that researchers can be \emph{committed} to a model-type -- have a favored model-type and focus on improving that model-type as a means to making research progress. Commitments to model-types also manifest in structuring machine learning workshops. Since in machine learning, workshops are meant to provide smaller venues to focus on making progress, this indicates that centering on a model-type can be seen as a way of making progress. For example, this year's International Conference on Machine Learning (ICML) included at least seven workshops focused specifically on deep learning\footnote{The deep learning workshops at 2019 ICML were: ``Theoretical Physics for Deep Learning"; ``Uncertainty and Robustness in Deep Learning"; ``Synthetic Realities: Deep Learning for Detecting AudioVisual Fakes"; ``Understanding and Improving Generalization in Deep Learning"; ``Identifying and Understanding Deep Learning Phenomena"; ``On-Device Machine Learning \& Compact Deep Neural Network Representations"; and ``Invertible Neural Networks and Normalizing Flows"}.

Researchers who are committed to a certain model-type think of that model-type as generalizable - that the success that the model-type had on some problems is an indication that, if we put more work into it, it will do well on many other problems. A commitment to a model-type is therefore fueled by some exemplars, some cases of great success for the model-type which are taken to be strong evidence for generalizability. For example, deep learning's success on a particular computer vision challenge, called ImageNet, was taken as evidence for its potential future success on other types of problems (we discuss ImageNet in greater detail in Section \ref{sec:imagenet}).

A commitment to a model-type has downstream effects for the research that is done. Model-types guide the research agenda of machine learning practitioners who are committed to them, and when many people are committed to the same model-type the discipline and the resources available to practitioners change. In particular, we argue that a commitment to model-types guides the selection of problems, constrains the creation of solutions, and promotes prerequisites around supporting tools and technologies.

\subsection{How a commitment to a model-type influences research}

\subsubsection{Problem selection}

Those who are committed to a model-type work to increase its precision and expand its scope. 

Different model-types are naturally good at different things. Those who are committed to a model-type work to improve its performance in those areas in which it is doing less well. For example, deep learning models have done very well in the field of computer vision and, more recently, in other fields such as natural language processing and reinforcement learning. However, deep learning has done less well in areas such as those involving causal, logical, or probabilistic reasoning. Other model-types, such as functional causal models, decision rules, or probabilistic graphical models (PGMs), are often better at these problems. But  researchers are now working to improve the ability of deep learning models to represent causal \citep{lopez2017discovering}, logical  \citep{cai2017making}, or probabilistic knowledge  \citep{kingma2013auto,rezende2014stochastic}.

In this way, commitments to model-types skew the selection of problems by virtue of the model-type's varying strengths and weakness.

\subsubsection{Constraining the search for solutions}

Model-types also help constrain the solutions considered to problems. When a research is committed to a model-type, she believes working on the model-type will be the most fruitful path to progress. Thus, she primarily pursues improvements to her committed model-type, rather than pursuing improvements to other model-types or devising a new model-type all-together. 

Improvements to a model-type are often made by revising \emph{hyperparameters}, which are the parameters which must be specified by the researcher rather than learned from the data. In deep learning, hyperparameters include specification of the mathematical transformations between layers, how many units are in each layer, how many layers there are, and so on. For example, an important hyperparameter innovation in deep learning was the change from using a sigmoidal or tanh function for the non-linear transformation between layers to using a ReLU function, a change that is often considered to have been essential to the revival of deep learning \citep{krizhevsky2012imagenet}.

\subsubsection{Prerequisites}

The success of a model-type depends on associated prerequisites around supporting tools, technologies, and information. A commitment to a model-type shapes the discipline by promoting and reinforcing such prerequisites. Consider deep learning as an example.

\begin{figure}[t]
    \centering
    \includegraphics[width=0.9\columnwidth]{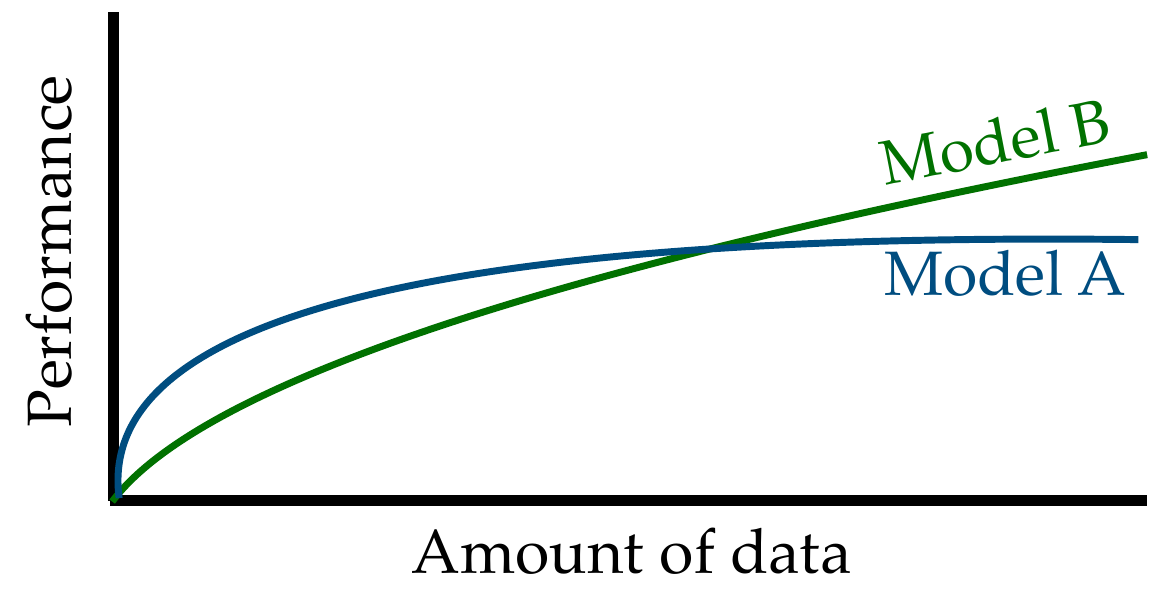}
    \label{fig:data}
    \caption{A toy rendition of two different ways that model performance can improve with data. Model A performs better than Model B in a low-data regimen, but Model B performs better in a high-data regime.}
    \label{fig:data-performance}
\end{figure}

One prerequisite for the success of a deep learning is the amount of data available. Machine learning algorithms are data-driven, and are supposed to get better at a given task the more data they are provided. However, there are different ways in which a model's performance may improve with more data. Figure \ref{fig:data-performance} depicts two such ways -- one model (curve A) has better performance with smaller amounts of data and the other model (curve B) has better performance with larger amounts of data. For any model, its ``data efficiency'', its performance as a function of the amount of data, will vary depending on the application. But generally, different model-types tend to achieve success with different levels of data. Deep learning typically requires large data sets to perform well \citep{hestness2017deep,sun2017revisiting}. Other methods, support vector machines, linear models, probabilistic graphical models, etc., can often do better on smaller data sets. Thus, deep learning requires a larger data prerequisite than other model types.

Another prerequisite for deep learning is extensive \emph{compute power}, the amount of computations that a computer can process within a given time frame. All model-types benefit from running on computers that have a lot of compute power, but some model-types benefit more than others. A similar graph to Figure \ref{fig:data-performance} can also hold with compute power on the $x$-axis. Similar to the data case, deep learning would follow the pattern of curve B. Unlike other model types, it performs better given access to large amounts of compute power. Thus, the availability of a lot of compute power may be a prerequisite for deep learning, but not for other model-types. 

Third, model-types require specialized technologies. For example, deep learning requires graphics processing units (GPUs), a specialized type of hardware. GPUs are useful for deep learning because they speed up the process of performing matrix multiplications, a computation that is essential to deep learning models but not necessarily for other model-types. Model-types can also benefit from specialized software that makes it easier for researchers to create models of that type. For example, deep learning benefited from the creation of software packages that automatically perform backpropagation, an algorithm used to compute derivatives that is primarily used to train deep learning models. Other model-types can also benefit from such software packages, but much less.

A commitment to a model-type involves promoting the prerequisites that model-type needs, for example building the right tools, collecting enough data, or buying the amount of compute resources necessary. When many people are committed to the same model-type, the resources available to the discipline at large change, reinforcing the predominance of that model-type. For example, as the popularity of deep learning increased, industry labs like Facebook and Google have put effort into developing easy-to-use supporting software, such as PyTorch \citep{paszke2017automatic} and Tensorflow \citep{abadi2016tensorflow}, which have greatly reduced the barrier to entry for creating deep learning models. Furthermore, many industry labs have also designed hardware to accelerate deep learning by creating components that are specialized for the computations used by deep learning models \citep{sze2017efficient}. For example, Nvidia frames one of their recent GPUs as ``Tesla P100: The Fastest Accelerator for Training Deep Neural Networks'' \citep{nvidia2016p100}.

The mass commitment to deep learning has promoted the availability of data, compute power, and the specialized technologies required for deep learning. 

\begin{figure*}[t]
    \centering
    \includegraphics[width=1.6\columnwidth]{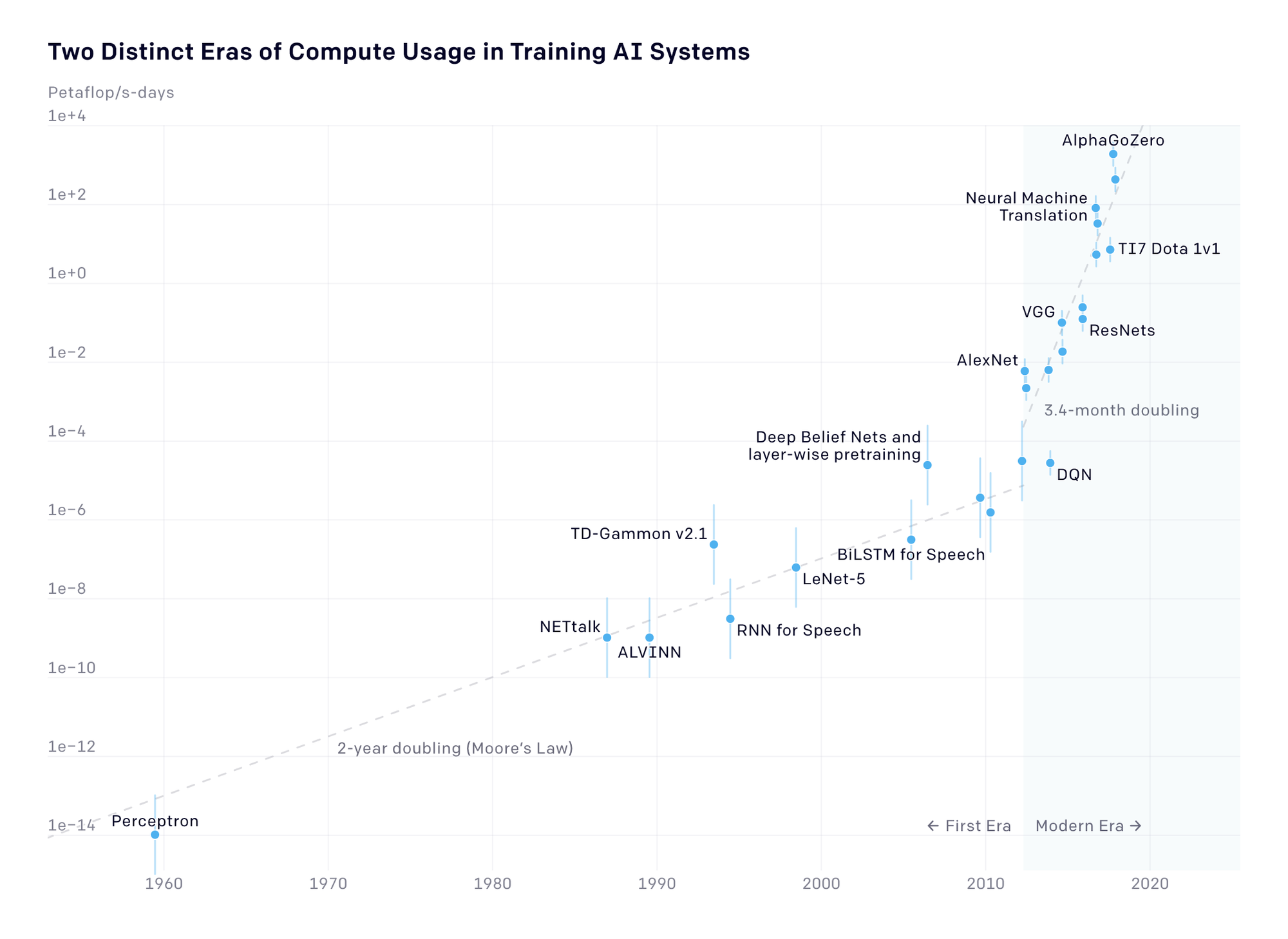}
    \caption{A figure (reproduced from OpenAI \citep{amodei_hernandez_2018}) depicting the change in amount of computational power used to train key results in machine learning. The amount of compute power is measured on the $y$-axis in petaflop/s-days, which translates to the number of operations (e.g. add, multiply) performed during model training divided by roughly $10^{20}$. Before 2012, the amount of compute power follows Moore's law and doubles at a 2-year rate. However, after 2012, the amount of compute power used increases much more rapidly -- doubling every 3.4 months. As we can see, 2012, the year that marked the rise of deep learning, also marked a drastic shift in the reliance on compute power.}
    \label{fig:compute}
\end{figure*}

\subsection{Similarities and differences between model-types and traditional concepts}

The functionality of model-types in machine learning is similar to the functionality of Kuhnian paradigms and Lakatosian research programmes in science. However, it is also different in important ways.

We argued that researchers who are committed to a model-type work to increase the scope and precision of that model-type.  This functionality is analogous to the way paradigms\footnote{In the narrowest sense, a Kuhnian paradigm is an exemplar - a widely accepted solution to a problem. See Hoyningen-Huene  (\citeyear{Hoyningen-Huene1993}) for more discussion of the different senses of "paradigm"} function in the natural sciences according to Kuhn (\citeyear{Kuhn1962}). Kuhn argues that most of the time scientists are concerned with increasing the scope and precision of their paradigm. Theoretical and experimental scientists are engaged in activities such as applying analogous solutions to more and more problems, getting better agreement between observation and theory, elaborating the theory to make it easier to compare it to observations, and so on (e.g. Kuhn, 1962, pp. 25-230). 

We have also argued that researchers who are committed to a model-type look for ways to overcome problems by making adjustments within the confines of that model-type. They may tune the hyperparameters (which are the aspects of the model that must be specified by the researcher). Or when it doesn't work, they may blame the prerequisites, e.g., in the case of deep learning, argue that there isn't sufficient compute power or data. This functionality is analogous to the way scientists work with research programmes according to Lakatos (\citeyear{lakatos1970falsification}). Lakatos argues that scientific theories are composed of core hypotheses, which are the hypotheses scientists mainly defend, and auxiliary hypotheses, which connect between the core hypotheses and observations. For example, in the Newtonian research programme, the core hypotheses are the three laws of motion. The auxiliary hypotheses include definitions of the terms the laws use (such as mass), procedures for measuring the quantities the laws are about (such as mass and velocity), methodologies on the correct use of scales, what reasonable margins of errors are, and so on. Upon conflicts with experience, scientists tend to revise the auxiliary hypotheses to save the core from falsification. When a prediction fails one can always blame the instruments, initial conditions, and so on. As a result of modifying the core hypotheses, a series of theories is created - that is the research programme. 

However, model-types differ from paradigms and research programmes in that they seem to rely much less on activities of theory-making. Model-types themselves are not theories, and are not composed of hypotheses like research programmes (at least not straightforwardly).\footnote{Some may wonder whether deep learning can be a theory -- a theory of how the brain works. Indeed, the earliest neural network (connectionist) models were inspired by neuroscience. However, to most modern machine learning researchers, who are more interested in the performance gains produced by deep learning, this is only a tangential connection. And as neuroscience has progressed, it has become clear that the neural networks used in machine learning are far simpler than the neural networks in the brain \citep{crick1989recent,barrett2019analyzing}.} Researchers may have hypotheses about which model-types will be the most successful, but the model-type itself does not seems to consist of theories or hypotheses in the same way that a paradigm or research programme does. 

The differences between the activities in machine learning and other disciplines invite not only naming new concepts (which is what we have done with ``model-type"), but also re-examination of these traditional philosophy of science concepts. For example: To what extent would it be useful to apply traditional concepts from philosophy of science, such as ``theory", ``paradigm", or ``research programme", to machine learning? How central are activities of theory-making to the these traditional concepts, given that it seems possible to reproduce their functionality without it? Thinking about such questions is among the ways in which machine learning can enrich philosophy of science.
\section{Comparison between model types is model-type-laden} \label{sec:imagenet}
How can model-types be compared? Some natural ways include comparing the problems that the model-type is successful at solving and the reasonableness of the background assumptions the model-typess rely on. However, as we have argued, a commitment to a model-type involves taking a stand on these very issues. Thus, being committed to a certain model-type means not being neutral with respect to the measures of success of model-types. We can't exactly say that a person committed to model-type A would become committed to model-type B if it is shown that model B solves the important problems better. Rather, changing your commitment may involve changing your views on which problems are important and what it means for a solution to be ``better".

In this section, we illustrate this complexity using the rise of deep learning.  We start by examining a common explanation for the rise of deep learning---the success of a deep learning model in the ImageNet competition.
 
\subsection{ImageNet}
ImageNet is a large-scale database of over 14 million images that was curated for the goal of furthering computer vision and related research \citep{deng2009imagenet}. Between 2010-2017, ImageNet ran an annual competition called  the ``ImageNet Large Scale Visual Recognition Challenge", which is also commonly referred to as simply ``ImageNet" \citep{russakovsky2015imagenet}. The task for the competition was to classify images into one of 1000 known classes. No deep learning models were submitted in the years 2010 and 2011, and the best error rate was 25.8\% \citep{lin2011large,sanchez2011high}. In 2012, the winner was the single deep learning model which was submitted, AlexNet \citep{krizhevsky2012imagenet}. AlexNet achieved a 16.4\% error rate, a remarkable 10\% percent lower than the runner-up.

This success is commonly cited as a trigger for the rise of deep learning. For example, Yann Lecun, Yoshua Bengio, and Geoffrey Hinton, who received the Turing Award in 2019 for their work in deep learning, said that \citep{lecun2015deep}:
\emph{``ConvNets [a type of neural network] were largely forsaken by the mainstream computer vision and machine-learning communities until the ImageNet competition in 2012.''}
 
By 2014, nearly all the submissions to the ImageNet challenge were deep learning models and the error rate steadily decreased. Deep learning became increasingly popular not only within the competition, but also outside of it --- in the industry at large as well as in academia. %
 
\subsection{Does ImageNet justify the rise of deep learning?}
In the natural sciences, success in a given experiment cannot justify a shift to a different paradigm or research programme on its own. It requires, among other things, prioritizing standards and preferences. To borrow Lakatos's example (\citeyear{lakatos1970falsification}), consider comparing Newton's early theory of optics, which focused on light-refraction, and Huyghens's early theory of light, which focused on light interference. We could compare between the two using experiments pertaining to, for example light-refraction. If we considered these experiments as crucial, then we would adopt Newton's theory over Huyghen's theory. But in doing so, we would also implicitly be elevating the problem of light-refraction over the problem of light interference. When prioritizing problems researchers are not necessarily under the illusion that, at the moment the prioritization is made, one theory is superior to all others in all respects. Rather, there is great hope and anticipation that the chosen theory's success on the puzzle of interest indicates success on other puzzles as well. The question is then: why do some experiments trigger this hope while others do not?

Similarly, in explaining the popularity of a model-type it is not enough to point out success in some competition. The question is not in which competitions a model-type did well, but rather why the success in some competitions rather than others had impact on the discipline. Deep learning did well in other competitions prior to ImageNet, but these didn't make nearly the same impact on the discipline as its success in ImageNet. For example, J\"urgen Schmidhuber's team at the Dalle Molle Institute for Artificial Intelligence Research, won four computer vision competitions with deep learning models \citep{ciresan2011flexible,ciresan2011flexible,ciresan2012deep} between May 15, 2011 and September 20, 2012 \citep{schmidhuber_2017}. The 2012 ImageNet competition was September 30, 2012. Matthew Zeiler, one of the winners of the 2013 ImageNet challenge, also notes that \citep{gershgorn_2017}: \textit{``This Imagenet 2012 event was definitely what triggered the big explosion of AI today. There were definitely some very promising results in speech recognition shortly before this... but they didn't take off publicly as much as that ImageNet win did in 2012 and the following years."} Furthermore, while deep learning models did very well on ImageNet and other computer vision and speech recognition competitions, they did less well in other areas, such as causal, probabilistic, or logical reasoning, again calling into question why ImageNet had such a large impact.

ImageNet, but not prior competitions, sparked great hope that deep learning could generalize in the field of computer vision and beyond. For example, Ilya Sutskever, one of the winners of the 2012 Imagenet challenge, said, \textit{``It was so clear that if you do a really good on ImageNet, you could solve image recognition''} \citep{gershgorn_2017}. Indeed, in a paper at CVPR 2019 (a computer vision conference) \citep{kornblith2019better} claim, \textit{"An implicit hypothesis in modern computer vision research is that models that perform better on ImageNet necessarily perform better on other vision tasks. However, this hypothesis has never been systematically tested."}  Furthermore, the influence of ImageNet spread outside of the field of computer visions, and triggered hopes of  generalization in other fields as well.

So what, if anything, makes ImageNet special? One important factor was that ImageNet's database was much larger. Compare ImageNet to its main predecessor, the PASCAL VOC challenge \citep{Everingham10,Everingham15}. PASCAL VOC was a better established image classification challenge, and in the first two years that the ImageNet challenge was hosted, it was co-located with PASCAL VOC challenge as a mere ``taster" competition. However, PASCAL VOC 2010 had only 20 classes and 19,737 images, while the ImageNet 2010 had 1000 classes and 1,461,406 images.

When the paper detailing the ImageNet database was originally published in 2009, skeptics disputed the value of such a large-scale database. Jia Deng, the lead author of the paper, said \citep{gershgorn_2017} \textit{``There were comments like 'If you can't even do one object well, why would you do thousands, or tens of thousands of objects?'"}. And yet, after ImageNet, many people take for granted the importance of large datasets.

Thus, the 2012 ImageNet challenge did not simply showcase the high performance of deep learning, it also marked a shift in how researchers thought progress would be made. More and more people began to believe that the field could make significant progress simply by scaling up datasets \citep{sun2017revisiting}. 

Furthermore, 2012 ImageNet also impacted the way people conceived of the role of compute power in progress. This is reflected in the increase in the amount of compute power used to train models after 2012. Researchers from OpenAI found that before 2012, the amount of compute power used to train neural networks was doubling at a 2-year rate in correspondence to Moore's law. Since 2012, when the current deep learning boom began, the amount of compute power used to train deep learning models that achieve state-of-the-art results has been increasing exponentially, doubling every 3.5 months \citep{amodei_hernandez_2018}. Both trends are shown in Figure \ref{fig:compute} (reproduced with permission from OpenAI).

Thus, the rise of deep learning came hand in hand with the rise of a new way to assess the success of models -- in data-rich and compute-rich environments.  Indeed, the shift is advocated for by the authors of the winning 2012 ImageNet model \citep{krizhevsky2012imagenet}: \textit{"All of our experiments suggest that our results
can be improved simply by waiting for faster GPUs and bigger datasets to become available."} 

It would be natural to assume that these shifts in the conception of progress were the result of new tools or resources that were not present before. For example, the increase in compute power was possible through the use of GPUs and techniques like GPU parallelization. However, the mere availability of these tools does not by itself imply that we \textit{ought} to evaluate progress in environments that are rich in data and compute power. Doing so assumes that we should define ``progress'' based on metrics that do not depend on the amount of data or compute power used, e.g. classification accuracy. However, we may have good reason to reject a data and compute-independent notion of progress. For example, we may reject to increased data collection for privacy reasons or the use of increased computational power for environmental costs. We will discuss this point in more detail in Section \ref{sec:values}.

In conclusion, we cannot straightforwardly say that ImageNet shows that deep learning is better than other model-types. This explanation takes for granted prerequisites of deep learning - that model-types should be evaluated in data-rich and compute-rich environments. If models are evaluated under the conditions in which deep learning models performs better, then the cards are stacked in favor of deep learning. The more general point here is that evaluation of model-types depends on considerations on which people committed to different model-types would disagree. Therefore, evaluation of model-types is model-type-laden: it depends on which model-type the evaluator is committed to.

\subsection{Incommensurability?}

The claim that comparison between model-types is model-type-laden is analogous to the claim that comparison between paradigms is paradigm-laden, i.e. that paradigms are incommensurable. Since incommesurability has been harshly criticized, one might wonder how plausible it is to claim that comparison between model-types is model-typle-laden. However, key criticisms to incommensurability are not applicable in the case of machine learning.   

What is incommensurability? In \emph{The Structure of Scientific Revolutions}\footnote{Kuhn dedicated much of his work after \emph{The Structure of Scientific Revolutions} to  developing the concept of incommensurability. While some think that his concept of incommensurability has changed greatly in subsequent work (e.g. \cite{Sankey1993}), others thinks the later work is a refinement of the earlier work (e.g. \cite{Hoyningen-Huene1993})} , Kuhn highlighted three interconnected aspects of incommensurability: semantic, ontological/perceptual, and methodological.\footnote{We follow the distinction made by Bird (\citeyear{Bird2018}). See \citep{Hoyningen-Huene1993,Sankey1993,Sankey2001} for different versions of this distinction.} From the semantic perspective, we can't say that the laws of one paradigm are derivable from the laws of a different paradigms due to the semantic differences between them. In other words, paradigms are not straightforwardly comparable because key terms don't mean or refer to the same things. For example, it is not the case that Newton's laws are derivable from Einsteinian mechanics because key terms, such as mass, don't mean the same thing. Second, we can't say that one paradigm describes the world better than another because the world itself is different for people working within different paradigms. The reason is that Kuhn argues that observations crucially depend on the theory held by those who take them, that is - that observation is theory-laden. Since the conceptual apparatus of different paradigms is different, the observations made by practitioners of different paradigms are different. So different that practitioners of different paradigms essentially practice science in different worlds. Third, we can't say that one paradigm is better than another because it does a better job at solving important problem. The reason is that, since the worlds of practitioners of different paradigms are different, the list of problems of interest are different. In addition, the standards of success are different. 

The semantic aspect of Kuhn's incommensurability received most of the critical attention, at least in philosophy of science \citep{Scheffler1982,Sankey2018,Mizrahi2018}.\footnote{The focus on the semantic aspect is perhaps due to the fact that Kuhn's later work (e.g. Kuhn, \citeyear{Kuhn1982}) as well as Feyerabend's (\citeyear{Feyerabend1978}) version of incommensurabily focus on semantic incommensurability. Other criticisms of incommensurability focus on the implications of incommensurability. See, e.g. Lauden (\citeyear{Lauden1996}) and \citeauthor{Gattei2003} (\citeyear{Gattei2003}).} However, the incommensurability-like effect in machine learning doesn't directly depend on semantics. Model-types are incommensurable because people who are committed to different model-types would disagree on the measures to use to compare them. For example, among other things, the dominance of deep learning involves shifting to favoring evaluations in data-rich and compute-rich environments. There is no need to appeal to the use of specialized languages or the theory-ladenness of observations to see this methodological incommensurability in machine learning. Moreover, semantic and ontological/perceptual incommensurability are less convincing than they are in the case of natural sciences. As researchers are not directly invested in theorizing about what the world is like, there is no need to think of the work of researchers working in different model-types as resulting in commitments to specialized language about what the world is like and shape observations. In other words, there is no need to think of researchers committed to different model-types as working in different worlds or unable to fully communicate about the world. Thus, methodological incommensurability in machine learning is independent from semantic and ontological/perceptual incommensurability.

\section{Comparison between model types is value-laden} \label{sec:values}
Having conceptualized how progress takes place in machine learning, we can now see how values shape the discipline at large. We have argued that prioritization of model-types is model-type-laden, in the sense that it depends on considerations on which people committed to different model-types would disagree. We now argue those same considerations are also value-laden in the sense that they implicitly encode political, social, and other values. Therefore, prioritization of model-types is not only model-type-laden but also value-laden. We illustrate on the case of deep learning.

\subsection{Prerequisites}

Prioritization of model-types requires favoring one set of prerequisites over another. When the prerequisites encode social and political values, the prioritization is value-laden. Let's consider two of the prerequisites for deep learning as examples.  

\subsubsection{Compute Power}

We have pointed out that, unlike other model-types, deep learning requires using a lot of compute power and a lot of GPUs in particular. This prerequisite is a carrier of a political value - centralization of power. Centralization of power, or rather decentralization of power, is already explicitly used for comparison between procedures and techniques in science and medicine. The general point is that tools that can only be made or used by a selected few, because they are complex or expansive, contribute to the concentration power. Such tools are only available to those with means, and they sustain and deepen the dependency between those who are in a position to provide the services and those who need those services \citep{Longino1995}. For example, in agriculture, sophisticated technologies create dependency on those who have the means and expertise to use them. Techniques that are accessible and can be locally implemented, such as small scale sustainable agriculture, promote decentralization of power. People who advocate for techniques of this sort are making the power dynamics involved in utilizing certain tools and procedures explicit and favor those which decentralize power.

Kevin Elliott (\citeyear{KevinC.Elliott2017}) highlights these issues with regards to the vitamin A deficiency crisis. Vitamin A deficiency is an acute problem among the poor worldwide, to the extent that hundreds of thousands of people become blind or die of it every year. One proposed course of action is to utilize a genetically modified species of rice, called "golden rice", which is enriched with vitamin A. Another alternative is to look into which of the crops that are indigenous to the relevant areas are rich in vitamin A, and encourage locals to consume them. Elliott argues that addressing the vitamin A problem using golden rice caters the western biochemical community, which not only stands to benefit from selling golden rice but also related tools of western agriculture that would likely accompany it such as pesticides and fertilizers (\citeyear{KevinC.Elliott2017}, p. 42).

The compute power prerequisite promotes centralization of power in two related ways. First, since GPUs are very expansive they create an entry barrier that favors those with financial means, such as big companies in rich countries. In addition, GPUs sustain and deepen the dependency on the major corporations that produce or can afford them. Hopefully, the cost of GPUs will drop substantially over time. However, like golden rice, the compute power prerequisite promotes centralization of power even if it is cheap. Even if golden rice is cheap, it creates a need to rely on products that would not be necessary using other techniques. 

The contribution of the popularization of deep learning to centralization of power has also been noticed by other researchers. For example, \citeauthor{strubell2019energy} (\citeyear{strubell2019energy}) note that: \textit{``Limiting this style of research to industry labs
hurts the NLP [natural language processing] research community in many ways. First, it stifles creativity. Researchers with good ideas but without access to large-scale compute will simply not be able to execute their ideas, instead constrained to focus on different problems. Second, it prohibits certain types of research on the basis of access to financial resources. This even more deeply promotes the already problematic ``rich get richer" cycle of research funding, where groups that are already successful and thus well-funded tend to receive more funding
due to their existing accomplishments. Third, the
prohibitive start-up cost of building in-house resources forces resource-poor groups to rely on cloud compute services such as AWS, Google Cloud and Microsoft Azure."}

The compute power prerequisite also encodes environmental values. Using environmental values to compare between tools and procedures in science and agriculture is not new. For example, one of the reasons Greenpeace objects to using golden rice is due to environmental concerns (Elliott, \citeyear{Elliott2017}, p 43). In machine learning, the extensive computational resources required by deep learning models take a toll on the environment. For example, \citeauthor{strubell2019energy} (\citeyear{strubell2019energy}) found that training one especially large state-of-the-art deep learning model resulted in carbon emissions that were over three times the amount of the carbon emissions in one car's average lifetime . These large environmental impacts have also been noticed by the Allen Institute for Artificial Intelligence, which recently released a position paper stating that ``Green AI" would be an emerging focus at the institute \citep{Schwartz2019GreenA}.

\subsubsection{Data}

Another prerequisite for deep learning is the availability of large data sets for training. Large data sets introduce several complexities. First, they create entry barriers because not everyone has access to sufficiently large amounts of data. This gives an advantage to large companies and promotes centralization of power, like the compute power prerequisite. 

Second, the collection of data about people, e.g. their location, heartbeat rates, or clicks, introduces an additional set of complexities around privacy, which bear on individual freedoms. For example, a 2009 House of Lords report on surveillance (\citeyear{house_of_lords_2009}) stated that \textit{``Mass surveillance has the potential to erode privacy. As privacy is an essential pre-requisite to the exercise of individual freedom, its erosion weakens the constitutional foundations on which democracy and good governance have traditionally been based in this country."} Deep learning requires mass collection of data and when this data is sensitive data about individuals, then it comes in tension with the value individual freedom which is associated with privacy.

Furthermore, it is a fact that we have more of certain kinds of data about some groups rather than others, and this data can be abused. For example, an anonymous programmer created a deep learning-based app called ``DeepNude", which allowed users to upload a photo of any woman and receive a fake, undressed version of the photo. The creator stated that the app only worked for women because, due to pornography, it is easier to find images of nude women online \citep{cole_2019}. Technology that requires large amounts of data to work furthers the power structures underlying what kinds of data are available about certain groups in the first place.

\subsection{Evaluation Criteria}
In comparing between theories, one needs to rely on characteristics of the theories as a whole - e.g. are they internally consistent? Are they consistent with established theories? Do they entail accurate predictions? Do they serve humanity well? Do they promote equal opportunity? Such characteristics of theories are often called ``theoretical virtues". Evaluating theories based on their theoretical virtues is a value-laden activity when theoretical virtues are carriers of values. 

Some of these virtues, such as applicability to human needs, wear their value commitments on their sleeves. However, even virtues that appear neutral, such as simplicity and consistency, are at least sometimes carriers of political, social, or other values (this was pointed out by, e.g., Kuhn (1977), and Longino (1996). To borrow two examples from Longino, consider consistency with established theories. If the established theories are sexist, new theories will also need to be sexist to be consistent with them. Similarly, simplicity favors theories with fewer kinds of entities. Therefore, theories in biology which treat all humans as versions of a man are simpler than theories which have multiple archetypes of humans. However, these simpler theories are androcentric. Moreover, even if virtues like simplicity and consistency are not themselves politically or socially loaded, comparing between them and virtues that are is loaded. Suppose theory A is simpler and more consistent with established theories (for whatever reason) and theory B is more applicable to human needs. Even if simplicity and consistency are not politically loaded, favoring theory A over theory B involves taking a stand on what it more important and that is loaded. 

Similar points apply to evaluation criteria in machine learning. Well known evaluation criteria for models include accuracy, explainability, transparency, and fairness. Some of these criteria, like fairness, are explicitly politically loaded. But even criteria that appear neutral on first glance may be carriers of values because of what it takes to satisfy them. Accuracy is one example. Which considerations is it permissible to use when attempting to make accurate predictions? For example, when is it permissible to use racial identity in making predictions about recidivism? Today, many think that we should not use social identity attributes, such as race and gender, to make such predictions even if they increase accuracy.\footnote{We note as a separate point that many in the machine learning and ethics community have pointed out that ``fairness through unawareness" is an insufficient solution because machine learning classifiers can still pick up on proxies for the sensitive attributes (\citeauthor{dwork2012fairness}, \citeyear{dwork2012fairness}; \citeauthor{hardt2016equality}, \citeyear{hardt2016equality}).}

The disapproval of using protected identities in making predictions is reflected in regulation. In the US, laws such as the Fair Housing Act (FHA), the Equal Credit Opportunity Act (ECOA), and the Fair Credit Reporting Act (FCRA), are being implemented on predictions based on big data and prohibit the use of protected identities for making decisions on loans, employment, and so on. For example, a 2016 report by the Federal Trade Commission (FTC)  determines that:
    ``an employer may not disfavor a particular protected group because big data analytics show that members of this protected group are more likely to quit their jobs within a five-year period. Similarly, a lender cannot refuse to lend to single persons or offer less favorable terms to them than married persons even if big data analytics show that single persons are less likely to repay loans than married persons." (\citeyear{federal2016big}, p. 18)
There are also restrictions on using generalizations based on proxies of protected identities, such as an address. For example, it is prohibited to deny a loan request because data analytics has found that people who live in the same zip code are generally not creditworthy (p. 16).

If one thinks that using these attributes is wrong, even if they yield accurate predictions, it is because some other value is more important than accuracy. For example, \citeauthor{Bolinger2018} (\citeyear{Bolinger2018}) argues that accepting  generalizations based on racial stereotypes is morally and epistemically wrong even if the stereotypes are statistically accurate \citep{Bolinger2018}. She gives the following example to illustrate this claim (originally from Gendler (\citeyear{Gendler2011})): John Hope Franklin is hosting an event to celebrate being awarded the Presidential Medal of Freedom. All other black men on the premises are uniformed attendants. Mistaking Franklin for an attendant, a woman hands him her coat check ticket and demands he brings her the coat. Statistically, since all other black men in the party are attendants, the woman's prediction that Franklin is an attendant is accurate. However, the assumption that Franklin is an attendant still feels wrong. Bolinger's explanation is that the problem is that the prediction is based on a racial stereotype (that black people have a lower social status), which is wrong even if the stereotype is statistically accurate. On Bolinger's view, relying on racial stereotypes is wrong because of cumulative effects. If the only time someone assumed Franklin has a low social status was at that party, the harm would have been minimal, just a one-off mistake based on a correct generalization. But when the same type of assumption is made consistently, as it is in the case of racial stereotypes, it interferes with black people's ability to signal authority and high social status. This results in limiting their opportunities for advancement which is incompatible with respecting their moral equality and autonomy.

Bolinger's explanation is of course only one attempt explain what it wrong with relying on generalizations based on social identity. Whatever the details are, the general point is that accuracy is not necessarily a value-neutral evaluation criterion. One reason is that decisions on which considerations are permissible to use in attempting to make accurate predictions are value-laden. For example, when predictions are accurate because they rely on racial generalizations, accuracy is a carrier of social values. The accuracy of algorithms that intentionally avoid relying on racial generalizations is also value-laden: in those cases accuracy is a carrier of the value of equality. Moreover, even putting aside the fact that accuracy is not value neutral, prioritizing accuracy over other virtues is not neutral because the other virtues are not. 

What these examples illustrate is that high accuracy is not the only thing that matters -- \emph{how} a model achieves high accuracy is also important. Therefore, interpertability, the ability to articulate why a model made a certain prediction, is in competition with accuracy. Deep learning models are often described as ``black-boxes'' because how they achieve high accuracy is difficult to scrutinize. Other model-types, such as graphical models, decision trees, or support vector machines, are typically easier to scrutinize to understand \emph{how} a decision was made. There are multiple ways to address the tension between accuracy and interpretability. A person who prioritizes accuracy may be more inclined to try to make deep learning algorithms more amenable to scrutiny \citep{simonyan2013deep,zeiler2014visualizing,olah2018building}. On the other hand, if interpretabililty is essential, then it may be more appealing to simply use a model that is already more amenable to interpretability.\footnote{Hybrid approaches, like ones that approximate a deep learning model with a simpler model-type, e.g. linear models, decision trees, are also popular (\citeauthor{ribeiro2016should}, \citeyear{ribeiro2016should}).} What we see here is that disagreements on which approach to take and which model-type to use also encode prioritizations on values. In this case, between accuracy and interpretability. Thus, the rise and fall of a model-type may encode the rise and fall of some values. In particular, increased interest in values like fairness, explainability, and interpretability may motivate favoring model-types other than deep learning.

\section{Conclusion, and values in deliberation}
There is a simple story to be told about disciplinary shifts in machine learning -- that progress is made by better models. But as we have seen, what is ``better'' cannot be a purely objective choice. First, because the rise of model-types comes hand in hand with rise of corresponding ways to evaluate model-types. Second, because comparisons between model types are value-laden, encoding social, political, and environmental values. 

It is important to explicitly reveal values that are shaping the discipline, so that these values can be examined and changed when deemed undesirable. Some in the community have already begun pushing for reforms of such disciplinary values. For example, the Allen Institute for Artificial Intelligence's recent ``Green AI'' paper \citep{Schwartz2019GreenA} advocates for increasing effort in ``environmentally friendly and inclusive'' AI research. They suggest to introduce environmentally positive incentives in the research community by creating norms around reporting measures of \emph{efficiency}, e.g. accuracy as a function of computational cost, rather than simply accuracy alone.

A related question inspired by these issues is who should make decisions in what values are furthered? Who gets to have a voice? In talking about selection of problems in science, \citeauthor{Kitcher2011} (\citeyear{Kitcher2011}) argues that all sides should have a say, including laypersons \citep{Kitcher2011}. A question for machine learning is: is the same true for machine learning? Who should have a say about which criteria are important in evaluating model-types? That is itself another value-laden question.
\section*{Acknowledgements}
For extensive feedback on this paper, we would like to thank Lara Buchak and Shamik Dasgupta. For comments and helpful discussion on early drafts of this paper, we would like to thank Morgan Ames, Kevin Baker, Julia Bursten, Christopher Grimsley, Elijah Mayfield, John Miller, Ludwig Schmidt, David Stump, and Will Sutherland.

Smitha was supported by the National Science
Foundation Graduate Research Fellowship Program under Grant No. DGE 1752814. Any opinions, findings, and conclusions or recommendations expressed in this material are those of the author(s) and do not necessarily reflect the views of the National Science Foundation.

\bibliographystyle{unsrtnat}
\bibliography{refs}

\end{document}